\documentclass{article}
\usepackage{ijcai22}

\usepackage{times}

\usepackage{soul}
\usepackage{url}
\usepackage[hidelinks]{hyperref}
\usepackage[utf8]{inputenc}
\usepackage[small]{caption}
\usepackage{graphicx}
\usepackage{amsmath}
\usepackage{booktabs}
\title{A survey of neural models for the automatic analysis of conversation: Towards a better integration of the social sciences.}
\author{
Chloé Clavel$^1$\footnote{Contact Author}\and
Matthieu Labeau$^1$\And
Justine Cassell$^{2,3}$\\
\affiliations
$^1$LTCI, Telecom-Paris, Institut Polytechnique de Paris, France\\
$^2$School of Computer Science, Carnegie Mellon University, USA \\
$^3$Inria Paris, France\\
\emails
\{chloe.clavel, matthieu.labeau\}@telecom-paris.fr,
justine@cs.cmu.edu
}

\begin{document}

\maketitle

\begin{abstract}
%TACL 10 pages.%This short example shows a contrived example on how to format the authors' information for {\it IJCAI--21 Proceedings} using \LaTeX{}.
Some exciting new approaches to neural architectures for the analysis of conversation have been introduced over the past couple of years. These include neural architectures for detecting emotion, dialogue acts, and sentiment polarity. They take advantage of some of the key attributes of contemporary machine learning, such as recurrent neural networks with attention mechanisms and transformer-based approaches. However, while the architectures themselves are extremely promising, the phenomena they have been applied to to date are but a small part of what makes conversation engaging. In this paper we survey these neural architectures and what they have been applied to. On the basis of the social science literature, we then describe what we believe to be the most fundamental and definitional feature of conversation, which is its co-construction over time by two or more interlocutors. We discuss how neural architectures of the sort surveyed could profitably be applied to these more fundamental aspects of conversation, and what this buys us in terms of a better analysis of conversation and even, in the longer term, a better way of generating conversation for a conversational system. 
\end{abstract}

\section{Introduction}
%Idée de l'article: présenter les différentes conditions d'une interaction humain-agent fluide et naturelle

%voir  le modele des summary de AAAI \url{https://ojs.aaai.org/index.php/AAAI/article/view/17770} et Empowering Conversational AI is a Trip to Mars: Progress and Future of Open Domain Human-Computer Dialogues)

% ou idee pour publication https://ieeexplore.ieee.org/stamp/stamp.jsp?tp=&arnumber=8416973
%Survey/Review Papers (Surveys). A Survey is a well-focused manuscript that puts recent progress into a broader perspective and accurately assesses the limits of existing theories. Surveys will typically be limited to 15 pages in IEEE two-column style format.

With the rise of call centers, chat applications and social media sites it has become essential to be able to automatically analyze human conversation.  
%Automatic analysis can flag problematic conversations, or radicalized members of social groups.  
Automatic analysis can flag problematic conversations for customer relationship management, or identify the nature of the collaboration within a student group for e-learning. Automatically analyzing conversations is also paramount for conversational systems by enabling them to provide the right answers at the right time in the right way.  
%These systems may one day be able to flag problematic conversations, or radicalized members of social groups. 
With this in mind, a new generation of neural architectures such as recurrent neural networks has arisen to detect conversational phenomena. These architectures have good potential for modeling the context of a speaker-turn and the conversation dynamics between speakers. They have shown good performance for turn-level prediction of dialog acts, and basic emotion/sentiment categories. %(\textit{e.g.}, in the task of Emotion Recognition in Conversations \cite{PoriaReview})

\begin{figure}[htbp]
  \includegraphics[width=0.48\textwidth]{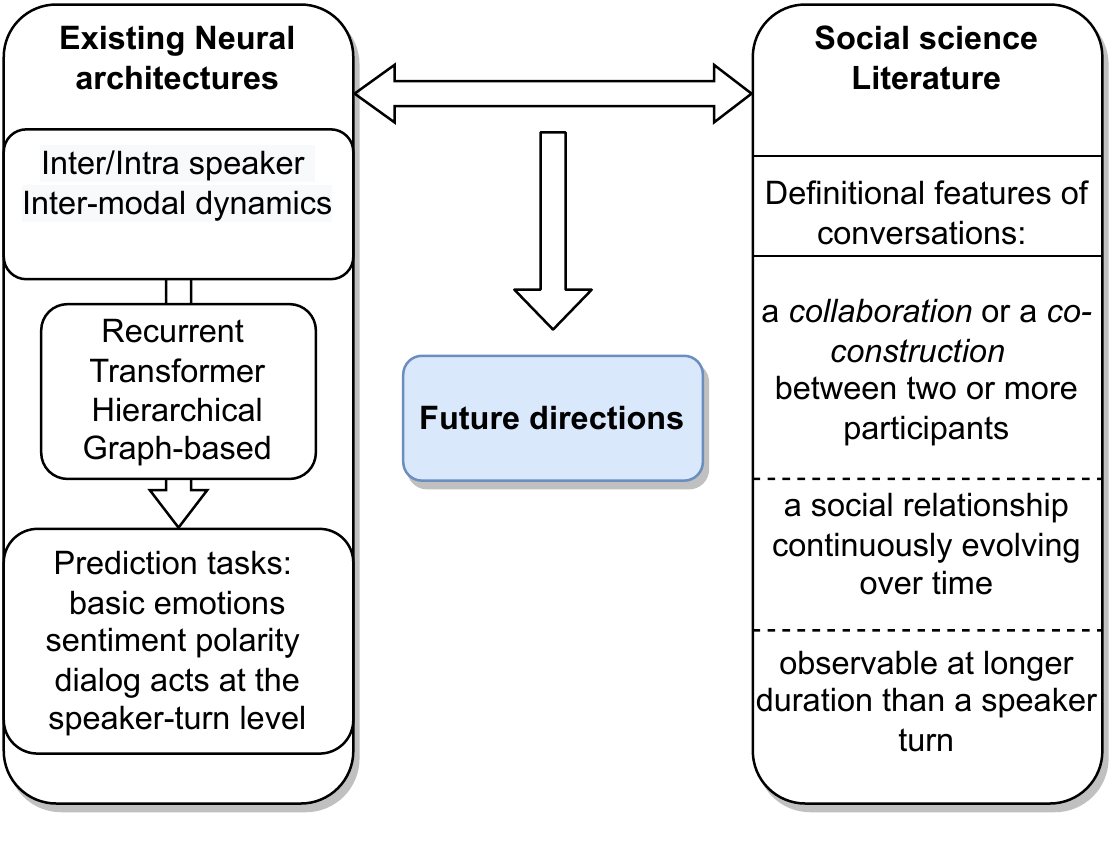}
  \caption{Overview of the paper structure: the confrontation of two fields of research}
  \label{fig:conversation}
\end{figure}

%While there exist some survey papers on the topic, such as \cite{PoriaReview,DengTAC}, they are outdated, and so in Section 2 we update them with the latest advances (\textit{e.g.}, transformer-based and graph-based).

More fundamentally, however, we argue that these neural architectures can be extended to other essential conversational phenomena, inspired by the social science literature. %Indeed, we argue that a key challenge is the ability of deep learning to join with previous social science research in human-human communication. 
In particular, when studying conversations, the social sciences have shown the importance of collaborative and dyadic processes in interactions and their joint and evolving mechanisms. But deep learning research has yet to be mobilized in this direction, despite calls to action such as \cite{KoppCoconstruction,eskenazi2020report}.  %Current neural architectures are thus modeling the speaker sequence of basic states  while ignoring the dyadic processes within an interaction and their joint and evolving mechanisms. 
In this paper, we take up the challenge by investigating how the recent neural architectures that we survey in Section 2 can be leveraged to mirror the collaborative and dyadic conversational processes highlighted by the social sciences, and that we survey in Section 3, along with some attempts to integrate those processes into computational systems. In Section 4 we then outline future directions for the deployment of neural architectures that take the social science perspective into account. In Section 5, we end the paper by opening the discussion on how these conversational features could also be integrated into current neural architectures for conversational systems in order for them to evolve into true partners to their human users.
%In this paper, we go further by investigating the capacities of recent neural architectures to mirror collaborative and dyadic processes as defined by the social sciences (Section 3), and by discussing how this definitional feature could also be integrated in current neural architectures for conversational systems in order for them to evolve into true partners to their human users (Section 4).

%1/commencer en parlant des nouvelles architectures et de quoi elles sont capables (traiter la dynamique interpersonnelle, integration du contetxe)  et quelles caracteristiques: 

%2/quels sont les phénomènes pour lesquels elles ont déjà été utilisées (emotion, senetiment, actes de dialogue)

%3/ Pourquoi ce n'est pas suffisant. a voir, la conversation est dyadique. Voila ce que c'est que le dyadique

%4/ comment les utiliser pour le dyadique

%\cite{clavel2015sentiment} 
While there exist survey papers on neural architectures for the analysis of conversation, such as \cite{PoriaReview,DengTAC}, they need to be supplemented with the latest advances (\textit{e.g.}, transformer-based and graph-based) and discussed from a social science perspective. The current paper thus constitutes the first survey in this area that comprehensively presents existing neural architectures and introduces related problems from a social science perspective, to provide research insights for the future. Given the short format of a survey paper, our coverage of current work is not exhaustive. For each type of neural architectures or each approach to dyadic processes, we give one representative example, knowing that other studies could also have been cited.

\section{Modeling Context in Conversations}

%Se positionner par rapport aux états de l’art récents: ***Conversational Machine Comprehension: A literature Review  \url{https://arxiv.org/abs/2006.00671}

%On ne va parler que des architectures neuronales mais on va parler du multimodal car c'est important pour les dyadic processes
%Pour le mulitmodal \url{https://ieeexplore.ieee.org/stamp/stamp.jsp?arnumber=9540240}
%\url{https://ieeexplore.ieee.org/abstract/document/9316758}

To date, the NLP community has concentrated on research gathered under the title of \textit{dialogue understanding}. This: primarily emotion recognition, sentiment analysis, dialogue intent classification, and dialogue state tracking. These tasks necessitate a transcript of conversational data, usually separated into utterances from different speakers, and mainly aim to classify the utterances using different possible sets of labels. In the following section, we focus on these tasks and how neural models are used to solve them, as a way of describing the current state of affairs, and we clarify what bridges can be built to the perspective we describe in Section~\ref{sec:socialscience}. %, as we believe they best reflect how NLP research is currently apprehending (the social component in dialog modeling [\textbf{TODO: à changer pour correspondre à l'introduction ?}]). %Of course, more complex, specific tasks exist, focused on particular phenomenons (\textit{e.g} sarcasm, humor). 
Conversation has a particular structure, with dependencies mirroring a number of interactions, as depicted in Figure~\ref{fig:conversation}: between speakers, but also among a speaker's own utterances~\cite{PoriaReview}. The relevant information may be close or distant, and may also be hidden behind a chain of co-references. Next, the labels associated with the aforementioned tasks are also interdependent. Finally, there exist dependencies among the different modalities involved in conversations, usually including natural language, nonverbal and acoustic features.  
Context, as in most of NLP, is thus key for dialogue: utterances are not considered independently, and neural models usually include neighboring utterances to make predictions. In order to extract contextual information, which is essential to understand conversational behavior, the current literature mainly leverages neural architectures. %Indicating the importance of this subject, a large empirical study attempting to unify previous approaches has recently been made~\cite{ghosal2020utterance}.

Specifically, the existing studies usually consider a conversation $\mathcal{C}$ as a sequence of utterances $(u_i)_{i=1}^{l}$. In order to map the sequence of utterances to a sequence of labels $(y_i)_{i=1}^{l}$, the trend is to use neural end-to-end models. In particular, the models usually follow the encoder-decoder paradigm, consisting of a feature extractor on which an auto-regressive language model is conditioned, and they use traditional architectures, such as recurrent, attention, transformers. In the present section, we focus on how and where
%, along this traditional division, 
context is integrated in addition to the representation of the current utterance. Our goal is to understand how current NLP tools -- mainly from the deep learning toolbox -- have shaped how context is used in dialogue, and what limits are associated to this use, which will allow us to better address them in the next sections. Hence, we will subsequently summarize recent progress in exploiting context for dialogue understanding tasks with sequential modeling; including recent transformer-based models making use of pre-training; then, with hierarchical modeling, followed by modeling dependencies between labels and, when available, between modalities.

\begin{figure}[t!]
  \includegraphics[width=0.48\textwidth]{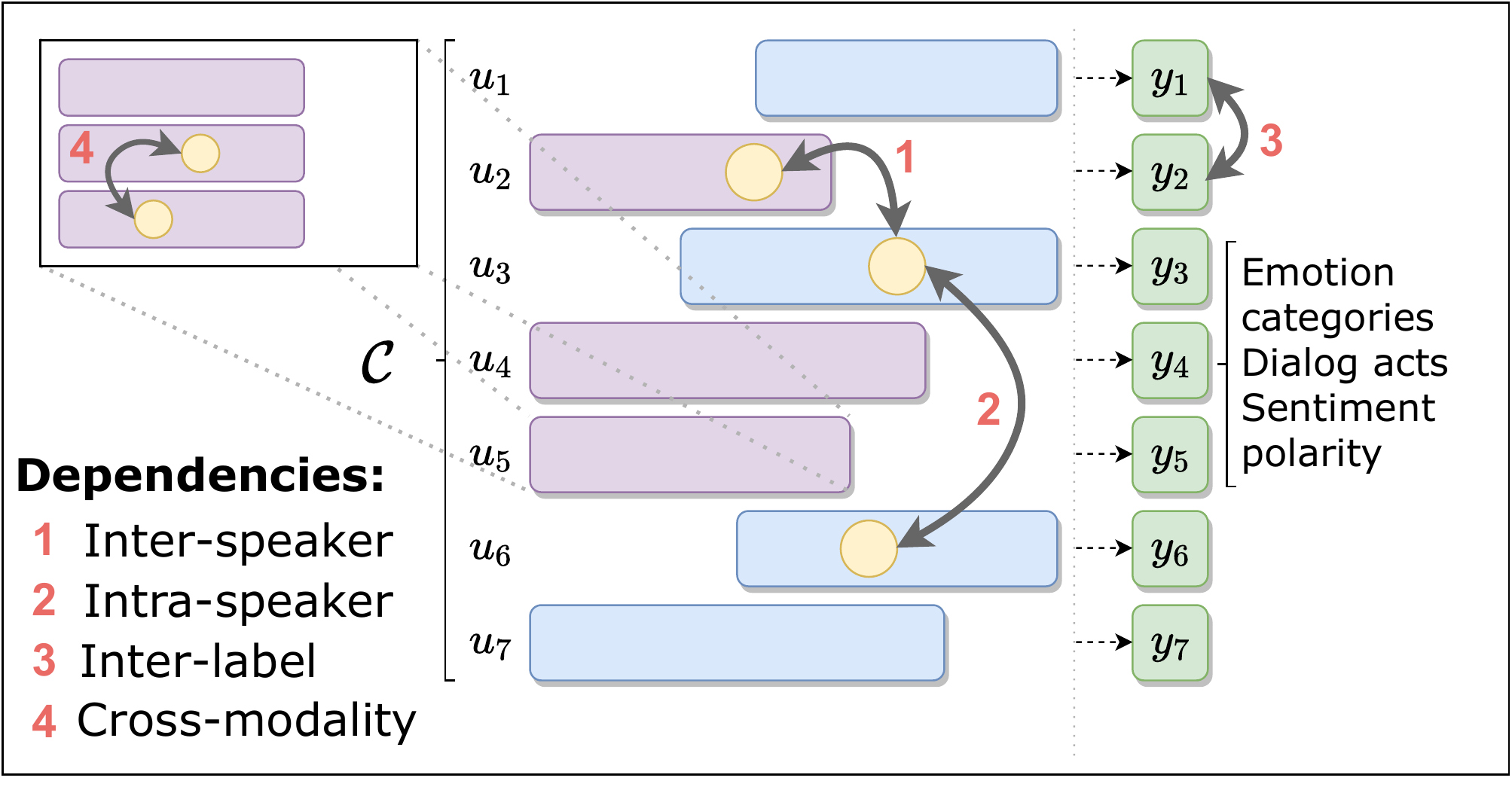}
  \caption{Conversation dynamics as modelled by current neural architectures}
  \label{fig:conversation}
\end{figure}

\subsection{Recurrent-based Sequence Modeling}

Since sequential architectures have become ubiquitous in NLP, it has become straightforward to encode an utterance into a representation of its words. In this context, capturing contextual dependencies between utterances has been done by carrying over the most recent utterance representation via memory~\cite{chen2016end-to-end}; then, past utterances without distinction~\cite{bapna2017sequential}. \cite{hazarika18conversational} were first to model self-speaker dynamics, using a recurrent architecture for each speaker. The extracted states are then fed to a recurrent module that models inter-speaker dynamics and maintains a global context of the conversation. \cite{majumder2019} improve on this by separating the speaker and listener into different roles, adding a recurrent architecture to track the global state of the conversation. \cite{xing20Adapted} propose a memory module to model interpersonal dependencies. Inversely, as a way to focus solely on emotion representations, \cite{lian2019domain} use adversarial training to obtain a common representation between speakers, building a speaker-independent model.

Recently, \cite{ghosal2020utterancelevel} studied the role of context for dialogue understanding tasks, presenting a series of exhaustive experiments with contextual baselines based on recurrent networks, among which the model of~\cite{majumder2019}, and a non-contextual one. Their first conclusion is that the contextual models perform largely better than the non-contextual in most cases; they also find out that the speaker-specific tracking of~\cite{majumder2019} is also useful in a large part of the experiments.

\subsection{Self-supervised Conversation Modeling}
\label{ssec:self-supervised}
The recent trend of using large transformer-based models like BERT~\cite{devlin-etal-2019-bert}, pre-trained on huge amounts of textual data, marks a milestone in context modeling: with specific \textit{self-supervised} and non-autoregressive learning objectives, they accumulate considerable knowledge about language. The most common of these objectives, Masked Language Modeling (MLM) and Next Sentence Prediction (NSP), have been adapted to the structure of dialogue, first with recurrent models~\cite{mehri-etal-2019-pretraining}, then with transformer-based models~\cite{chapuis2020hierarchical}; notably, by acting at both the word and utterance level (for example, by masking and retrieving complete utterances), these objectives encourage representations to integrate both local and global context. %written human-human interactions (corpus MultiWoz) Differences : nous BERT, eux encoder LSTM. Points communs : pre-training tasks similaires) %TODO ici identifier ce qui reste à faire pour que ce soit fait pour les interactions humain-agent  ou meme plus specifiquement pour l'analyse des comportements sociaux?
Such architectures are also used to build contextual representations for response selection~\cite{whang2020domain} %jiao-etal-2020-exploiting
and generation~\cite{zhang2019dialogpt} in dialogue and question answering systems: in Section~\ref{ssec:convAI}, we will come back to how these models are considered in conversational AI. 

Several studies investigate the representations they learn, whether based on recurrent or transformer networks:~\cite{sankar2019neural} show that they are not significantly affected by perturbations to the dialogue history. Similarly,~\cite{saleh2020probing} probe internal representations of dialogue models with several tasks, among which are dialogue understanding tasks, and find that shuffling the training data does not significantly decrease performance on probing tasks, even for model pre-trained with self-supervised objectives: hence, these architectures fail to leverage the turn-taking structure of dialogue. 
% In the discussion, we will come back to look into common points and differences with how these models are considered in the human-agent interaction literature. %addressing the wider range of system that are designed as conversational AIs. \textbf{[TODO: vérifier qu'on le fait bien.]}
\subsection{Hierarchical and Graph-based Conversation Modeling}

Most of the works mentioned until now model dialogue structure in some way, usually through encoding the difference between speakers - but we are interested here in how to model a conversation in structures more complex than flat sequences of utterances, which is often done with hierarchical neural architectures, or graphs. The hierarchical recurrent neural network in \cite{li2019dual} models the natural hierarchical structure of a conversation (character, word, utterance, and conversation levels), while \cite{chapuis2020hierarchical} proposes to pre-train hierarchical transformers on two levels (word and utterances). Very recently, \cite{hazarika2021conversational} presented an hybrid approach with a transformer at the word level and RNN at the utterance level, while~\cite{hu-etal-2021-dialoguecrn} propose a model that uses attention to iteratively extract emotional information from two global context representations modeling inter and intra-speaker dynamics. 

In~\cite{ghosal2019graph}%zhang2019modeling
, a conversation is modeled as a graph whose nodes and edges respectively represent utterances and temporal dependencies between them, and emotion recognition is modeled as a node classification task using graph convolutional networks (GCNs). These architectures allow addressing context propagation issues usually associated with RNNs on long sequences. \cite{ishiwatari-etal-2020-relation} propose to add positional information to the graph to take in account the classical sequential structure. Similarly,~\cite{shen-etal-2021-directed} use directed acyclic graphs to model the structure within the conversation, as an attempt to combine the strength of graph-based and recurrent-based approaches. But another issue is that conversational topics are not necessarily organized in a sequence during an interaction, which is why learning to refer to the relevant part of the context is important. %For dialog state tracking, \cite{zhong-etal-2018-global} uses global and local recurrent networks and self-attention to learn representations from not only the context, but past actions, while \cite{sharma-etal-2019-improving} re-use the same architecture to encode all available information and learn in which part of the past utterances to look.
For dialogue act classification,~\cite{wang-etal-2020-integrating} groups previous contexts by labels and integrate them to the representation through an heterogeneous graph, before using a GCN to model the interaction. Besides integrating the context through an interaction graph,~\cite{li-etal-2021-past-present} adds outside knowledge from a pre-trained common sense model to more accurately model relations between speakers. 

However, to the best of our knowledge, no experimental study looks further than performance on dialogue understanding tasks to confirm that hierarchical and graph-based conversational models are effectively leveraging the particular structure of dialogue, such as done for recurrent architectures.

%**TODO ref a ajouter pSychological-Knowledge-Aware Interaction Graph (SKAIG) des connaissances externes sur les interactions psychologiques entre les utterances sont intégrées au graphe. %https://underline.io/lecture/38585-past,-present,-and-future-conversational-emotion-recognition-through-structural-modeling-of-psychological-knowledge

\subsection{Modeling at the Decoder Level}

For a sequence labeling task, it is natural to try to model dependencies between neighboring labels: for example, with Conditional Random Fields (CRFs) on top of a neural recurrent model~\cite{Kumar2018DialogueAS}. In~\cite{colombo2020guiding}, the authors go further and compare the task of sequence labelling as Neural Machine Translation (NMT) which maps a sequence of words in a given language to a sequence of words in an another language leveraging progress made in NMT as most models generating text based on a separate textual input. In particular, with the popularization of the sequence-to-sequence architecture, sequence labeling tasks have begun to be treated as text generation, with the sequential decoder integrating the conversational aspects and contextual dependencies between the labels. Then, decoding can be guided by attention, while still enhanced with a supplementary CRF~\cite{colombo2020guiding}. \cite{lu-etal-2020-iterative} use an iterative emotion interaction network to model the interaction among emotion labels and deal with the lack of gold labels at inference time. Another way to leverage dependencies between labels is to train a model simultaneously on two different tasks, as~\cite{li2019dual}, who use attention and CRFs on two separate but linked decoders to predict both dialogue acts and topics. Similarly, \cite{qin2020dcrnet} propose 'relation layers' between encoder and decoder to model the interactions between emotions and dialogue states to be predicted by two separates decoders. 

\subsection{Aligning multiple modalities}

The shift to data-driven neural end-to-end systems allows one to automatically obtain a data representation, avoiding feature engineering. This means that these systems allow information coming from different modalities to be represented in the same space. A wide array of work has investigated how best to fuse these modalities for dialogue understanding tasks, going from simply concatenating features to proposing recurrent neural models capable of modeling both inter and intra-modality dynamics~\cite{zadeh-etal-2017-tensor}%Majumder18Multimodal
. However, modalities are heterogeneous, and often unaligned. Additionally, sequence-based networks have difficulty modeling long-range dependencies between modalities. The attention mechanism is a first step to address the issue~\cite{ghosal-etal-2018-contextual}, improved upon by using transformer models~\cite{tsai-etal-2019-multimodal}. Going further,~\cite{tang-etal-2021-ctfn} propose a translation-inspired fusion network, establishing pairwise translation modules between modalities with the purpose of dealing with missing modalities at prediction time.

\section{A Perspective from the Social Sciences}
\label{sec:socialscience}

While the architectures presented in the previous section are extremely promising, the phenomena they have been applied to to date represent but a small part of what makes conversation engaging - in fact, they don't take into account to the extent possible what defines conversation as studied by the social science literature. Before introducing the future directions, we look back at the main studies in social sciences defining what we believe to be the most fundamental and definitional feature of conversation, which is its co-construction over time by two or more interlocutors, and at the computational studies that aim to integrate such a feature for the automatic analysis of conversations.

%Changing the unit of analysis and the considered user's state critique des labels et unité d’analyse (modifier structure temporelle et structure conceptuelle)

\subsection{Some Definitions} Much of computer science, including some computational linguistics, has maintained a certain loyalty to Shannon and Weaver's 1948 Mathematical Theory of Communication \cite{shannon1948mathematical} whereby a message is passed from sender to receiver via a channel. While this model was extremely influential in its time, it has since been replaced in much of the social sciences by models that take into account the collaborative nature of conversation. For example, \cite{clark1996using} highlights the joint responsibility of the participants to ensure that contributions to the conversation are mutually understood, what he calls "grounding." In the ethnomethodology  or conversational analysis literature, \cite{schegloff2007sequence} highlights the contributions of the participants to the temporal sequentiality of talk, and how it can be moved forward via units such as response pairs where one interlocutor's contribution is not complete until the second interlocutor responds to it. That conversation evolves over time implies that, rather than looking at context as a function of one single utterance or turn, the analysis of these phenomena must include looking at time as continuous, and longer than a single turn or two turns - indeed, even stretching over multiple interactions.

Similarly, sociolinguists such as Spencer-Oatey \cite{SpencerOatey2005} highlight the joint contribution of interlocutors to the evolving social relationship, where one cannot even distinguish the contribution of each interlocutor, because the very concept of a social phenomenon such as rapport is dyadic or mutual and jointly evolves across the interaction  \cite{cassell2003negotiated}. In addition to rapport, other examples of dyadic phenomena joint between two or more speakers include alignment, affiliation \cite{stivers2008stance}, and interpersonal trust \cite{bickmore2001relational}. 
From both a social science and a computational perspective, the analysis of these phenomena implies a unit of analysis that is not an interlocutor or a message but a dyad or a group. Novel coding schemes, statistical analyses \cite{kenny2020dyadic}, and computational models all must be adapted to take into account the dyadic nature of these phenomena. We can contrast these phenomena with engagement, for example, that has a fairly long history in systems that analyze user behavior, but that does not need to be dyadic - one interactant can be engaged while the other is not. Likewise one participant in a conversation can be hostile towards the other without that second person being hostile - or even  aware of hostility.

 \subsection{Existing Coding Schemes}

Work integrating dyadic processes is still rare in deep learning for NLP, although there does exist research that develops coding schemes to provide ground truth for dyadic phenomena. One example comes from the annotation of how rapport evolves dyadically in  conversation \cite{zhao2014towards}, \cite{cerekovic2016rapport}; another comes from annotation of group cohesion %by act4team \cite{kauffeld201821} used 
in \cite{kantharaju2020multimodal} . The units used for the annotation reflect the different temporalities of dyadic processes. Both rely on fixed-size segments (annotation of rapport on windows from 30 seconds to 1-minute and, for cohesion, on 2-minute windows). \cite{ambady1992thin} have shown that the annotation of these very short windows demonstrates good test-retest reliability for a number of psychological processes, including rapport, and has high inter-rater reliability among 3 ore more annotators. 

In \cite{langlet2015improving}, on the other hand, the coding scheme relies on a smaller and turn-based unit, the adjacency pair, for annotation of shared likes and dislikes of the interaction partners. An additional way to obtain ground truth for dyadic processes, but at the level of the interaction, is to use self-reported measures obtained through questionnaires in order to evaluate models of an agent's overall nonverbal behavior (e.g., the rapport questionnaire in \cite{gratch2007creating}, \cite{cerekovic2016rapport}). It is not entirely clear, however, how reliable self-report is, nor whether the processes are truly dyadic, since only the human participants annotate their rapport.

\subsection{Existing Prediction Models}
Research also exists that provides automatic measures of dyadic processes. \cite{langlet2015improving} use reasoning models that take the form of hand-crafted linguistic rules for analyzing shared likes and dislikes of the interaction partners. Temporal association rules built from data mining are used in \cite{madaio2017using} for rapport estimation. %or \cite{janssoone2016using} for social stance analysis)
Classical machine learning models such as Support Vector Machines in \cite{muller2018detecting,hagad2011predicting} or regression models in \cite{cerekovic2016rapport} are used for rapport estimation.  Recurrent neural architectures such as bi-directional LSTM with temporal selective attention in  \cite{yu2017temporally} are used to integrate multimodal context. %or the engagement of a user in a human-robot interaction \cite{atamna2020hri}. 
 % and in \cite{atamna2020hri} and Temporally Selective Attention Model for Social and Affective State Recognition in Multimedia Content)and Gated Recurrent Units in \cite{atamna2020hri}. . 
 For the last two, the model relies strongly on the design of knowledge-driven multimodal features that allow social science knowledge about dyadic processes to be integrated.  In  
\cite{muller2018detecting}, the nonverbal rapport features  are speech, facial expressions, hand motions, and cross-modal synchronization features. Postural cues are studied in \cite{hagad2011predicting}. %in order to build a computational model of rapport based on various machine learning models (SVMs, MLP and Naive Bayes).  
%Regression models and classification models are studied in \cite{cerekovic2016rapport} to estimate rapport in human-agent interactions (Semaine dataset) using 
In \cite{cerekovic2016rapport}, multimodal features are mixed with information on human personality traits. The time window for the extraction of features is linked to that used for the supervision. The specificity of dyadic processes is to require the design of both intermodal and interspeaker synchronization features (e.g., the number of synchronous smiles in a given window).

The above mentioned work is representative of the majority of the studies carried out on automatically measured dyadic processes in interactions, i.e. they make very little use of the verbal modality and, methodologically, prefer knowledge-driven methods to deep learning methods.  %The reason for this preference is that in this context it is more complicated to access annotated data for deep learning frameworks and there is a great need for transparency in the functioning of perception and action mechanisms of conversational agents. %De plus, les méthodes d'apprentissage profond sont connues pour leur opacité.
Thus, neural models integrating dyadic processes are still rare in NLP research. %empathy?

\section{Future Directions}
\label{ssec:NNDyadic}

%\subsubsection{Existing work for neural architectures for Dyadic process measuring}
As described in Section 2, current neural architectures are able to model intra-speaker, inter-speaker and inter-modal dynamics. % (sequential models with attention mechanisms such as seq2seq and their hierarchical stacking \cite{colombo2020guiding}, graph neural networks \cite{ghosal2019graph}, or transformer-based approaches \cite{chapuis2020hierarchical})   
But what is still missing is the integration of the dyadic processes as a joint action of participants to understand each other, to maintain the flow of the interaction and to create a social relationship. 

\paragraph{Reframing the Supervision}  The various neural architectures seen in Section 2 
can be leveraged with a different kind of supervision than currently (\textit{e.g.}, using the speaker turn as a unit  and basic emotion categories as a label). It implies using units of analysis and labels of the kind proposed by the coding schemes presented in Section 3.2. (\textit{e.g}., a rapport level for each 30-second frame), and working on the last layers of the neural architectures in order to integrate the slowly evolving nature of co-construction processes. % that are slowly evolving along the interaction.

\paragraph{Use Intermediate Predictions} In dyadic processes, the response pair (that is, the unit of analysis that includes one speaker's contribution and the other's reply) can be taken as an intermediate level of analysis. One example comes from the automatic prediction of conversational strategies across speakers with the aim to improve the prediction of rapport \cite{Zhao2016}.  Another example comes from the automatic prediction of shared likes and dislikes in two speakers \cite{langlet2015improving}. An interesting direction is the training of the models to predict dyadic processes at these two levels (the response pair and the longer window) using joint learning objectives inspired by the recent advances in multi-task learning in NLP \cite{garcia-etal-2019-token}.

%\subsubsection{Dealing with small labelled datasets}
\paragraph{Developing New Datasets} Using such neural models is based on the assumption that there is enough annotated data in dyadic processes to build deep architectures. If large datasets of interactions are already available, only a small part of them are annotated for dyadic processes. Researchers in deep learning must mobilize to develop new datasets annotated for dyadic processes in collaboration with researchers from the social sciences.

\paragraph{Pre-training Objectives on Unlabelled Data} 
%Inject the formalism into self-supervised objectives at the encoder level for representing user states related to SODP. La deuxième limite est de trouver les bonnes représentations. 
  %in deep learning and natural language processing research, less attention has been given to adapt data-driven textual representations for dialogue until recently. 
As seen in Section 2, the integration of the conversational aspects by capturing contextual dependencies between utterances can be done through  unsupervised pretraining objectives (\textit{e.g.}, next utterance retrieval) for learning dialogue context representation using sequential encoders or transformer-based approaches%*(Chapuis et al., 2020)
. But the self-supervised objectives developed to date  are still rare and they do not exploit the full structural richness of dyadic processes.  
We need thus to develop new methods able to train representations on unlabelled data with pre-training objectives and architectures dedicated to the modelling of the definitional features of dyadic processes. The evaluation of what is exactly learned by these representations need to be done via dedicated methods such as probing (see Section~\ref{ssec:self-supervised}). We need also to work on a grounding of these evaluation methods in social sciences.

\paragraph{Tractable Models}
In order to handle the lack of labelled data, deep learning research offers method such as meta-learning, few-shot learning and multi-task learning in order to foster the tractability of the models \cite{deng2020low}. Such methods investigate how we can train models with few annotated data by relying on models trained on different data source with different labels. The application of such methods for the analysis of conversations is emerging and not trivial \cite{guibon-etal-2021-shot}.  We should leverage this research progress in order to take advantage of all the already existing corpora where similar but different phenomena (\textit{e.g.}, engagement, sentiment) are annotated.% Again, the existing relationships between the different tasks identified in T1.1 (e.g. engagement, rapport, affective trust) will be crucial to develop the meta-learning strategies. 

\paragraph{Hybrid Models} One promising approach, beginning to be found in the literature, is to integrate social science-based rules as features in the encoding stage of neural architectures. Similarly pre-trained representations using a knowledge base from the social science literature can be integrated into graph neural architectures \cite{li-etal-2021-past-present}. In addition to being particularly applicable to dyadic features, this approach has the benefit of allowing greater explanability of the output. It has, however, been little attempted to date.

 \section{The Case of Conversational AI}
\label{ssec:convAI}
As described in the introduction, the analysis of dyadic processes in conversation is essential to making conversational AI into a true partner to its human users. The goal here is different from that of the automatic analysis of conversation, as the system must \emph{generate} the utterances of one of the partners (the agent) \cite{clavel2015sentiment}. The question, then, lies in how to select or generate the agent’s utterances as a function of different criteria of relevance to the user's previous utterances. In this section, we point towards the future by examining how the extent to which existing work on neural conversational AI is conducive to the integration of dyadic processes.  %in order to foster the user’s propensity to socially collaborate with the agent.The agent utterances are seen as a response to user ones 
 %\subsubsection{A better coupling of tasks}
%We argue that we need to go further and efforts must be made in order to better build the supervision or obtain finer models of SODP as presented in Section 3.3. % We need to investigate new ways of integrating Natural Language Understanding into dialog strategies by proposing neural models that integrate user-agent interactional dynamics and that are able to foster SODP. 
 % models used for generating the agent’s social behavior and models used for detecting user’s state are considered independently or are combined in a very simple manner. %From the point of view of human-agent interaction systems, the measuring of the social dynamics from the user’s side enriches the user model at the input of the agent's interaction strategies in order to improve the relevance of the latter's response. Even more, it can be used to measure the impact of the agent's response on the user by analyzing the next user’s utterances. Both analyzing user utterances and generating the agent’s answer are thus interwoven. **ADDREF**
%the measurement and fostering of the user's propensity to socially collaborate through the agent's dialogue policy is a crucial, yet underexplored objective, of deep learning and NLP research for conversational systems.

\paragraph{Dyadic processes in modular approaches} Traditionally  the generation phase of conversational systems is carried out in a modular way. The agent's utterance is selected or generated according to a dialogue policy which is, in turn, selected according to the current user's state and intents. The three modules (tracking user’s state, selecting agent’s policy, and conditional generation of agent’s utterance) rely on separate neural architectures. In order to better take into account the dyadic processes of conversation using modular approaches, the user's state analysis module must be replaced by a module capable of analyzing the behaviors of the dyad formed by the agent and the user, considering them as interaction partners in the joint construction of the conversation. The selection of the agent's policy and the generation of the agent's utterance has to be chosen with this aim of fostering dyadic processes. Perhaps the most promising approach is to use reinforcement learning with a reward based on the relevant social science literature. This is in line with what has been proposed recently by \cite{pecune2020framework}. 

\paragraph{Dyadic processes in end-to-end systems} Recent work in the field of natural language generation for dialogue systems has moved to a reliance on end-to-end systems~\cite{gao2019neural}. By dispensing with the steps of classifying the user's utterance and identifying the relevant dialogue policy, end-to-end systems make it possible to directly generate  (such as by using DialoGPT \cite{zhang2019dialogpt}) or select responses (that is, to output the highest-ranking response to a question from a predetermined list) as a function of the preceding sequence of utterances in a dialogue (e.g., using memory networks in \cite{bordes2016learning}). Part of the power of these models is that they don't require any supervision. Such architectures have not been designed to integrate dyadic processes, but they can be a way to free the process from the turn-based analysis of a user's state implied by modular approaches. %**TODO a clarifier avec Matthieu In order to do generation with a GPT-2 model[Reference], \cite{cao-etal-2020-pretrained} study how to properly fuse different sources of information (the dialog history, the current reply, and the persona of the speaker) with various attention mechanisms. 
However, very recently, \cite{benotti2021grounding} calls into question the ability of these models that are simply trained on large amounts of successful dialogues to reproduce human-analogous collaborative grounding. Enabling end-to-end neural models of this sort to integrate dyadic processes without any strong supervision does not appear to be trivial. One approach is to leverage DialogGPT \cite{liu2021towards} or BlenderBot \cite{smith-etal-2020-put} but to introduce supervision into the generation process. For example, in \cite{liu2021towards},  DialoGPT is used with supervision to generate the agent's response as a condition of conversational strategies oriented towards the user's emotional support.  %Un premier exemple de ce qui peut être fait est donné dans article emotional support ACL 2021 qui s'appuie sur les state of the art generative dialog models DialoGPT pour générer des réponses conditionnellement à une stratégie de conversation dédié au support émotionnel de l'utilisateur. ** to add reference (ex. conditional sentence generation method such as the one we previously developed for style transfer (Colombo et al., submitted)) à l'état de l'art sur le conditional sentence generation for generating the agent’s answer in natural language given the selected dialog policy?  ** to add voir comment la stratégie est déterminée dans l'article
\cite{zhong2020towards} %
rely on retrieval-based conversational models (\textit{e.g.}, selecting the most relevant utterance for the agent from a set of utterances) and learn a neural model (Bert model named CoBERT) on the basis of empathetic conversations for response selection. The supervision stage here is relatively simple and relies on the selection of empathetic conversations to train the empathy model which is then compared to a non-empathetic model trained on non-empathetic conversations.  %We should investigate methods for learning the task of choosing the utterance among those given in the selected dialog policy. %** est-ce que c'est dans cet article ou la backpropagation est utilisé pour selectionner la dialog policy?non
In that approach, however, there is no option to explicitly control the level or the kind of empathy at the utterance level, and it is difficult to explain the rationale for the system's answer.

\paragraph{Open Discussion}  We argue that a smarter and multi-level coupling of analysis and generation is required in order to better integrate the dyadic processes of conversation. We have given examples of how modular approaches can control the generation process in such a way as to handle dyadic phenomena. It is less obvious how to integrate these dyadic processes without adding supervision in end-to-end generation models. %An important question is thus related to the relevance of end-to-end approaches in general for future socially-oriented operational dialog systems. %This requires the models to be able to implicitly integrate the status of the social relationship between the human and the agent in an interaction without the need for supervision either to measure it or to generate a socially-relevant answer.  %It seems thus difficult that les systèmes end-to-end soient capables d'apprendre de manière non supervisée les entrelacemements entre les énoncés de l'humain et ceux de l'agent, de saisir et répondre aux comportements que l'humain met naturellement en jeu face à la machine pour créer une relation sociale afin que l'interaction soit la plus efficace possible, ou encore que le système de dialogue soit lui-même capable de mettre en place ce genre de stratégies pour améliorer l'efficacité par rapport à la tâche.
Additionally, fully data-driven neural dialogue models raise the issue of how to control the nature of the input, an issue highlighted by systems such as Tay, that quickly learned to imitate the racist and sexist content its users served up \cite{eskenazi2020report}. Generating utterances for dialogue systems that distribute information across modalities, and where this information responds to the multimodal context of the dyad, also remains an open problem for unsupervised neural models for conversational AI.

\section{Conclusions}
Research in neural architectures for the analysis of conversation is booming, and has led to high-performing and innovative models. In parallel, although with a longer history, the social science study of conversation has provided an increasingly rich and relevant literature on the fundamental aspects of conversation.
The two research areas, however, are still quite disconnected, which deprives the computational analysis of conversation of some important insights and tools. After an inter-disciplinary literature review, the current paper identifies future directions that arise from cross-fertilization between the two research approaches.

In order for the cross-fertilization to take root, a common formalism will be required to make a bridge between structural aspects of conversation as defined by the social sciences and the architectures underlying the different neural models. As an example, we must progress beyond the prediction of a sequence of emotion categories by considering interaction between two or more participants (whether they are agents, people or a combination of the two) as a co-construction process evolving over time. We showed here some ways in which  interpersonal dynamics are already integrated into existing neural architectures. Now the architectures must be adapted in order to go beyond interpersonal modeling to considering the dyad formed by the participants as a single analytic unit.

\bibliographystyle{named}
\bibliography{IJCAI}

\begin{thebibliography}{}

\bibitem[\protect\citeauthoryear{Ambady and Rosenthal}{1992}]{ambady1992thin}
Nalini Ambady and Robert Rosenthal.
\newblock Thin slices of expressive behavior as predictors of interpersonal
  consequences: A meta-analysis.
\newblock {\em Psychological bulletin}, 111(2):256, 1992.

\bibitem[\protect\citeauthoryear{Bapna \bgroup \em et al.\egroup
  }{2017}]{bapna2017sequential}
Ankur Bapna, Gokhan T{\"u}r, Dilek Hakkani-T{\"u}r, and Larry Heck.
\newblock Sequential dialogue context modeling for spoken language
  understanding.
\newblock In {\em ACL}, 2017.

\bibitem[\protect\citeauthoryear{Benotti and
  Blackburn}{2021}]{benotti2021grounding}
Luciana Benotti and Patrick Blackburn.
\newblock Grounding as a collaborative process.
\newblock In {\em EACL}, 2021.

\bibitem[\protect\citeauthoryear{Bickmore and
  Cassell}{2001}]{bickmore2001relational}
Timothy Bickmore and Justine Cassell.
\newblock Relational agents: a model and implementation of building user trust.
\newblock In {\em SIGCHI}, 2001.

\bibitem[\protect\citeauthoryear{Bordes \bgroup \em et al.\egroup
  }{2016}]{bordes2016learning}
Antoine Bordes, Y-Lan Boureau, and Jason Weston.
\newblock Learning end-to-end goal-oriented dialog.
\newblock {\em arXiv preprint arXiv:1605.07683}, 2016.

\bibitem[\protect\citeauthoryear{Cassell and
  Bickmore}{2003}]{cassell2003negotiated}
Justine Cassell and Timothy Bickmore.
\newblock Negotiated collusion: Modeling social language and its relationship
  effects in intelligent agents.
\newblock {\em User modeling and user-adapted interaction}, 13(1):89--132,
  2003.

\bibitem[\protect\citeauthoryear{Cerekovic \bgroup \em et al.\egroup
  }{2016}]{cerekovic2016rapport}
Aleksandra Cerekovic, Oya Aran, and Daniel Gatica-Perez.
\newblock Rapport with virtual agents: What do human social cues and
  personality explain?
\newblock {\em IEEE Transactions on Affective Computing}, 8(3):382--395, 2016.

\bibitem[\protect\citeauthoryear{Chapuis \bgroup \em et al.\egroup
  }{2020}]{chapuis2020hierarchical}
Emile Chapuis, Pierre Colombo, Matteo Manica, Matthieu Labeau, and Chloe
  Clavel.
\newblock Hierarchical pre-training for sequence labelling in spoken dialog.
\newblock In {\em Findings of EMNLP}, 2020.

\bibitem[\protect\citeauthoryear{Chen \bgroup \em et al.\egroup
  }{2016}]{chen2016end-to-end}
Yun-Nung~Vivian Chen, Dilek Hakkani-Tür, Gokhan Tur, Jianfeng Gao, and
  Li~Deng.
\newblock End-to-end memory networks with knowledge carryover for multi-turn
  spoken language understanding.
\newblock In {\em Interspeech}, 2016.

\bibitem[\protect\citeauthoryear{Clark}{1996}]{clark1996using}
Herbert~H Clark.
\newblock {\em Using language}.
\newblock Cambridge university press, 1996.

\bibitem[\protect\citeauthoryear{Clavel and
  Callejas}{2015}]{clavel2015sentiment}
Chloe Clavel and Zoraida Callejas.
\newblock Sentiment analysis: from opinion mining to human-agent interaction.
\newblock {\em IEEE Trans. on Aff. Comp.}, 7(1), 2015.

\bibitem[\protect\citeauthoryear{Colombo \bgroup \em et al.\egroup
  }{2020}]{colombo2020guiding}
Pierre Colombo, Emile Chapuis, Matteo Manica, Emmanuel Vignon, Giovanna Varni,
  and Chloe Clavel.
\newblock Guiding attention in sequence-to-sequence models for dialogue act
  prediction.
\newblock In {\em AAAI}, 2020.

\bibitem[\protect\citeauthoryear{Deng and Ren}{2021}]{DengTAC}
Jiawen Deng and Fuji Ren.
\newblock A survey of textual emotion recognition and its challenges.
\newblock {\em IEEE Transactions on Affective Computing}, pages 1--1, 2021.

\bibitem[\protect\citeauthoryear{Deng \bgroup \em et al.\egroup
  }{2020}]{deng2020low}
Shumin Deng, Ningyu Zhang, Zhanlin Sun, Jiaoyan Chen, and Huajun Chen.
\newblock When low resource nlp meets unsupervised language model:
  Meta-pretraining then meta-learning for few-shot text classification.
\newblock In {\em AAAI}, 2020.

\bibitem[\protect\citeauthoryear{Devlin \bgroup \em et al.\egroup
  }{2019}]{devlin-etal-2019-bert}
Jacob Devlin, Ming-Wei Chang, Kenton Lee, and Kristina Toutanova.
\newblock {BERT}: Pre-training of deep bidirectional transformers for language
  understanding.
\newblock In {\em ACL-HLT}, 2019.

\bibitem[\protect\citeauthoryear{Eskenazi and Zhao}{2020}]{eskenazi2020report}
Maxine Eskenazi and Tiancheng Zhao.
\newblock Report from the nsf future directions workshop, toward user-oriented
  agents: Research directions and challenges.
\newblock {\em arXiv preprint arXiv:2006.06026}, 2020.

\bibitem[\protect\citeauthoryear{Gao \bgroup \em et al.\egroup
  }{2019}]{gao2019neural}
Jianfeng Gao, Michel Galley, and Lihong Li.
\newblock Neural approaches to conversational ai.
\newblock {\em arXiv:1809.08267}, 2019.

\bibitem[\protect\citeauthoryear{Garcia \bgroup \em et al.\egroup
  }{2019}]{garcia-etal-2019-token}
Alexandre Garcia, Pierre Colombo, Florence d{'}Alch{\'e} Buc, Slim Essid, and
  Chlo{\'e} Clavel.
\newblock From the token to the review: A hierarchical multimodal approach to
  opinion mining.
\newblock In {\em EMNLP-IJCNLP}, 2019.

\bibitem[\protect\citeauthoryear{Ghosal \bgroup \em et al.\egroup
  }{2018}]{ghosal-etal-2018-contextual}
Deepanway Ghosal, Md~Shad Akhtar, Dushyant Chauhan, Soujanya Poria, Asif Ekbal,
  and Pushpak Bhattacharyya.
\newblock Contextual inter-modal attention for multi-modal sentiment analysis.
\newblock In {\em EMNLP}, 2018.

\bibitem[\protect\citeauthoryear{Ghosal \bgroup \em et al.\egroup
  }{2019}]{ghosal2019graph}
Deepanway Ghosal, Navonil Majumder, Soujanya Poria, Niyati Chhaya, and
  Alexander Gelbukh.
\newblock {DialogueGCN: A Graph Convolutional Neural Network for Emotion
  Recognition in Conversation}.
\newblock In {\em EMNLP}, 2019.

\bibitem[\protect\citeauthoryear{Ghosal \bgroup \em et al.\egroup
  }{2020}]{ghosal2020utterancelevel}
Deepanway Ghosal, Navonil Majumder, Rada Mihalcea, and Soujanya Poria.
\newblock Utterance-level dialogue understanding: An empirical study.
\newblock {\em arXiv}, 2009.13902, 2020.

\bibitem[\protect\citeauthoryear{Gratch \bgroup \em et al.\egroup
  }{2007}]{gratch2007creating}
Jonathan Gratch, Ning Wang, Jillian Gerten, Edward Fast, and Robin Duffy.
\newblock Creating rapport with virtual agents.
\newblock In {\em IVA}, 2007.

\bibitem[\protect\citeauthoryear{Guibon \bgroup \em et al.\egroup
  }{2021}]{guibon-etal-2021-shot}
Ga{\"e}l Guibon, Matthieu Labeau, H{\'e}l{\`e}ne Flamein, Luce Lefeuvre, and
  Chlo{\'e} Clavel.
\newblock Few-shot emotion recognition in conversation with sequential
  prototypical networks.
\newblock In {\em EMNLP}, 2021.

\bibitem[\protect\citeauthoryear{Hagad \bgroup \em et al.\egroup
  }{2011}]{hagad2011predicting}
Juan~Lorenzo Hagad, Roberto Legaspi, Masayuki Numao, and Merlin Suarez.
\newblock Predicting levels of rapport in dyadic interactions through automatic
  detection of posture and posture congruence.
\newblock In {\em PASSAT, SocialCom}, 2011.

\bibitem[\protect\citeauthoryear{Hazarika \bgroup \em et al.\egroup
  }{2018}]{hazarika18conversational}
Devamanyu Hazarika, Soujanya Poria, Amir Zadeh, Erik Cambria, Louis-Philippe
  Morency, and Roger Zimmermann.
\newblock {Conversational Memory Network for Emotion Recognition in Dyadic
  Dialogue Videos}.
\newblock In {\em NAACL}, 2018.

\bibitem[\protect\citeauthoryear{Hazarika \bgroup \em et al.\egroup
  }{2021}]{hazarika2021conversational}
Devamanyu Hazarika, Soujanya Poria, Roger Zimmermann, and Rada Mihalcea.
\newblock Conversational transfer learning for emotion recognition.
\newblock {\em Information Fusion}, 65:1 -- 12, 2021.

\bibitem[\protect\citeauthoryear{Hu \bgroup \em et al.\egroup
  }{2021}]{hu-etal-2021-dialoguecrn}
Dou Hu, Lingwei Wei, and Xiaoyong Huai.
\newblock {D}ialogue{CRN}: Contextual reasoning networks for emotion
  recognition in conversations.
\newblock In {\em ACL-IJCNLP}, 2021.

\bibitem[\protect\citeauthoryear{Ishiwatari \bgroup \em et al.\egroup
  }{2020}]{ishiwatari-etal-2020-relation}
Taichi Ishiwatari, Yuki Yasuda, Taro Miyazaki, and Jun Goto.
\newblock Relation-aware graph attention networks with relational position
  encodings for emotion recognition in conversations.
\newblock In {\em EMNLP}, 2020.

\bibitem[\protect\citeauthoryear{Kantharaju \bgroup \em et al.\egroup
  }{2020}]{kantharaju2020multimodal}
Reshmashree~B Kantharaju, Caroline Langlet, Mukesh Barange, Chlo{\'e} Clavel,
  and Catherine Pelachaud.
\newblock Multimodal analysis of cohesion in multi-party interactions.
\newblock In {\em LREC}, 2020.

\bibitem[\protect\citeauthoryear{Kenny \bgroup \em et al.\egroup
  }{2020}]{kenny2020dyadic}
David~A Kenny, Deborah~A Kashy, and William~L Cook.
\newblock {\em Dyadic data analysis}.
\newblock Guilford Publications, 2020.

\bibitem[\protect\citeauthoryear{Kopp and Krämer}{2021}]{KoppCoconstruction}
Stefan Kopp and Nicole Krämer.
\newblock Revisiting human-agent communication: The importance of joint
  co-construction and understanding mental states.
\newblock {\em Frontiers in Psychology}, 12, 2021.

\bibitem[\protect\citeauthoryear{Kumar \bgroup \em et al.\egroup
  }{2018}]{Kumar2018DialogueAS}
Harshit Kumar, Arvind Agarwal, Riddhiman Dasgupta, and Sachindra Joshi.
\newblock Dialogue act sequence labeling using hierarchical encoder with crf.
\newblock In {\em AAAI}, 2018.

\bibitem[\protect\citeauthoryear{Langlet and
  Clavel}{2015}]{langlet2015improving}
Caroline Langlet and Chlo{\'e} Clavel.
\newblock Improving social relationships in face-to-face human-agent
  interactions: when the agent wants to know user’s likes and dislikes.
\newblock In {\em ACL-IJCNLP}, 2015.

\bibitem[\protect\citeauthoryear{Li \bgroup \em et al.\egroup
  }{2019}]{li2019dual}
Ruizhe Li, Chenghua Lin, Matthew Collinson, Xiao Li, and Guanyi Chen.
\newblock A dual-attention hierarchical recurrent neural network for dialogue
  act classification.
\newblock In {\em CoNLL}, 2019.

\bibitem[\protect\citeauthoryear{Li \bgroup \em et al.\egroup
  }{2021}]{li-etal-2021-past-present}
Jiangnan Li, Zheng Lin, Peng Fu, and Weiping Wang.
\newblock Past, present, and future: Conversational emotion recognition through
  structural modeling of psychological knowledge.
\newblock In {\em Findings EMNLP}, 2021.

\bibitem[\protect\citeauthoryear{Lian \bgroup \em et al.\egroup
  }{2019}]{lian2019domain}
Zheng Lian, Jianhua Tao, Bin Liu, and Jian Huang.
\newblock Domain adversarial learning for emotion recognition, 2019.

\bibitem[\protect\citeauthoryear{Liu \bgroup \em et al.\egroup
  }{2021}]{liu2021towards}
Siyang Liu, Chujie Zheng, Orianna Demasi, Sahand Sabour, Yu~Li, Zhou Yu, Yong
  Jiang, and Minlie Huang.
\newblock Towards emotional support dialog systems.
\newblock {\em arXiv preprint arXiv:2106.01144}, 2021.

\bibitem[\protect\citeauthoryear{Lu \bgroup \em et al.\egroup
  }{2020}]{lu-etal-2020-iterative}
Xin Lu, Yanyan Zhao, Yang Wu, Yijian Tian, Huipeng Chen, and Bing Qin.
\newblock An iterative emotion interaction network for emotion recognition in
  conversations.
\newblock In {\em COLING}, 2020.

\bibitem[\protect\citeauthoryear{Madaio \bgroup \em et al.\egroup
  }{2017}]{madaio2017using}
Michael Madaio, Rae Lasko, Amy Ogan, and Justine Cassell.
\newblock Using temporal association rule mining to predict dyadic rapport in
  peer tutoring.
\newblock {\em International Educational Data Mining Society}, 2017.

\bibitem[\protect\citeauthoryear{Majumder \bgroup \em et al.\egroup
  }{2019}]{majumder2019}
Navonil Majumder, Soujanya Poria, Devamanyu Hazarika, Rada Mihalcea,
  Alexander~F. Gelbukh, and Erik Cambria.
\newblock {DialogueRNN: An Attentive {RNN} for Emotion Detection in
  Conversations}.
\newblock In {\em AAAI}, 2019.

\bibitem[\protect\citeauthoryear{Mehri \bgroup \em et al.\egroup
  }{2019}]{mehri-etal-2019-pretraining}
Shikib Mehri, Evgeniia Razumovskaia, Tiancheng Zhao, and Maxine Eskenazi.
\newblock Pretraining methods for dialog context representation learning.
\newblock In {\em ACL}, 2019.

\bibitem[\protect\citeauthoryear{M{\"u}ller \bgroup \em et al.\egroup
  }{2018}]{muller2018detecting}
Philipp M{\"u}ller, Michael~Xuelin Huang, and Andreas Bulling.
\newblock Detecting low rapport during natural interactions in small groups
  from non-verbal behaviour.
\newblock In {\em IUI}, 2018.

\bibitem[\protect\citeauthoryear{Pecune and
  Marsella}{2020}]{pecune2020framework}
Florian Pecune and Stacy Marsella.
\newblock A framework to co-optimize task and social dialogue policies using
  reinforcement learning.
\newblock In {\em IVA}, 2020.

\bibitem[\protect\citeauthoryear{Poria \bgroup \em et al.\egroup
  }{2019}]{PoriaReview}
Soujanya Poria, Navonil Majumder, Rada Mihalcea, and Eduard Hovy.
\newblock Emotion recognition in conversation: Research challenges, datasets,
  and recent advances.
\newblock {\em IEEE Access}, 7:100943--100953, 2019.

\bibitem[\protect\citeauthoryear{Qin \bgroup \em et al.\egroup
  }{2020}]{qin2020dcrnet}
Libo Qin, Wanxiang Che, Yangming Li, Minheng Ni, and Ting Liu.
\newblock Dcr-net: A deep co-interactive relation network for joint dialog act
  recognition and sentiment classification.
\newblock {\em arxiv/2008.06914}, 2020.

\bibitem[\protect\citeauthoryear{Saleh \bgroup \em et al.\egroup
  }{2020}]{saleh2020probing}
Abdelrhman Saleh, Tovly Deutsch, Stephen Casper, Yonatan Belinkov, and Stuart
  Shieber.
\newblock Probing neural dialog models for conversational understanding.
\newblock In {\em Workshop ACL on NLP for Conversational AI}, 2020.

\bibitem[\protect\citeauthoryear{Sankar \bgroup \em et al.\egroup
  }{2019}]{sankar2019neural}
Chinnadhurai Sankar, Sandeep Subramanian, Christopher Pal, Sarath Chandar, and
  Yoshua Bengio.
\newblock Do neural dialog systems use the conversation history effectively? an
  empirical study.
\newblock {\em arXiv.1906.01603}, 2019.

\bibitem[\protect\citeauthoryear{Schegloff}{2007}]{schegloff2007sequence}
Emanuel~A Schegloff.
\newblock {\em Sequence organization in interaction: A primer in conversation
  analysis I}.
\newblock Cambridge university press, 2007.

\bibitem[\protect\citeauthoryear{Shannon}{1948}]{shannon1948mathematical}
Claude~Elwood Shannon.
\newblock A mathematical theory of communication.
\newblock {\em The Bell system technical journal}, 27(3):379--423, 1948.

\bibitem[\protect\citeauthoryear{Shen \bgroup \em et al.\egroup
  }{2021}]{shen-etal-2021-directed}
Weizhou Shen, Siyue Wu, Yunyi Yang, and Xiaojun Quan.
\newblock Directed acyclic graph network for conversational emotion
  recognition.
\newblock In {\em ACL-IJCNLP}, 2021.

\bibitem[\protect\citeauthoryear{Smith \bgroup \em et al.\egroup
  }{2020}]{smith-etal-2020-put}
Eric~Michael Smith, Mary Williamson, Kurt Shuster, Jason Weston, and Y-Lan
  Boureau.
\newblock Can you put it all together: Evaluating conversational agents{'}
  ability to blend skills.
\newblock In {\em ACL}, 2020.

\bibitem[\protect\citeauthoryear{Spencer-Oatey}{2005}]{SpencerOatey2005}
Helen Spencer-Oatey.
\newblock (im)politeness, face and perceptions of rapport: Unpackaging their
  bases and interrelationships.
\newblock 1(1):95--119, 2005.

\bibitem[\protect\citeauthoryear{Stivers}{2008}]{stivers2008stance}
Tanya Stivers.
\newblock Stance, alignment, and affiliation during storytelling: When nodding
  is a token of affiliation.
\newblock {\em Research on language and social interaction}, 41(1):31--57,
  2008.

\bibitem[\protect\citeauthoryear{Tang \bgroup \em et al.\egroup
  }{2021}]{tang-etal-2021-ctfn}
Jiajia Tang, Kang Li, Xuanyu Jin, Andrzej Cichocki, Qibin Zhao, and Wanzeng
  Kong.
\newblock {CTFN}: Hierarchical learning for multimodal sentiment analysis using
  coupled-translation fusion network.
\newblock In {\em ACL-IJCNLP}, 2021.

\bibitem[\protect\citeauthoryear{Tsai \bgroup \em et al.\egroup
  }{2019}]{tsai-etal-2019-multimodal}
Yao-Hung~Hubert Tsai, Shaojie Bai, Paul~Pu Liang, J.~Zico Kolter,
  Louis-Philippe Morency, and Ruslan Salakhutdinov.
\newblock Multimodal transformer for unaligned multimodal language sequences.
\newblock In {\em Proceedings of the 57th Annual Meeting of the Association for
  Computational Linguistics}, 2019.

\bibitem[\protect\citeauthoryear{Wang \bgroup \em et al.\egroup
  }{2020}]{wang-etal-2020-integrating}
Dong Wang, Ziran Li, Haitao Zheng, and Ying Shen.
\newblock Integrating user history into heterogeneous graph for dialogue act
  recognition.
\newblock In {\em COLING}, 2020.

\bibitem[\protect\citeauthoryear{Whang \bgroup \em et al.\egroup
  }{2020}]{whang2020domain}
Taesun Whang, Dongyub Lee, Chanhee Lee, Kisu Yang, Dongsuk Oh, and HeuiSeok
  Lim.
\newblock An effective domain adaptive post-training method for bert in
  response selection.
\newblock In {\em Interspeech}, 2020.

\bibitem[\protect\citeauthoryear{{Xing} \bgroup \em et al.\egroup
  }{2020}]{xing20Adapted}
S.~{Xing}, S.~{Mai}, and H.~{Hu}.
\newblock Adapted dynamic memory network for emotion recognition in
  conversation.
\newblock {\em IEEE Trans. on Aff. Comp.}, pages 1--1, 2020.

\bibitem[\protect\citeauthoryear{Yu \bgroup \em et al.\egroup
  }{2017}]{yu2017temporally}
Hongliang Yu, Liangke Gui, Michael Madaio, Amy Ogan, Justine Cassell, and
  Louis-Philippe Morency.
\newblock Temporally selective attention model for social and affective state
  recognition in multimedia content.
\newblock In {\em ACM Multimedia}, 2017.

\bibitem[\protect\citeauthoryear{Zadeh \bgroup \em et al.\egroup
  }{2017}]{zadeh-etal-2017-tensor}
Amir Zadeh, Minghai Chen, Soujanya Poria, Erik Cambria, and Louis-Philippe
  Morency.
\newblock Tensor fusion network for multimodal sentiment analysis.
\newblock In {\em EMNLP}, 2017.

\bibitem[\protect\citeauthoryear{Zhang \bgroup \em et al.\egroup
  }{2019}]{zhang2019dialogpt}
Yizhe Zhang, Siqi Sun, Michel Galley, Yen-Chun Chen, Chris Brockett, Xiang Gao,
  Jianfeng Gao, Jingjing Liu, and Bill Dolan.
\newblock Dialogpt: Large-scale generative pre-training for conversational
  response generation.
\newblock {\em arXiv preprint arXiv:1911.00536}, 2019.

\bibitem[\protect\citeauthoryear{Zhao \bgroup \em et al.\egroup
  }{2014}]{zhao2014towards}
Ran Zhao, Alexandros Papangelis, and Justine Cassell.
\newblock Towards a dyadic computational model of rapport management for
  human-virtual agent interaction.
\newblock In {\em International Conference on Intelligent Virtual Agents},
  pages 514--527. Springer, 2014.

\bibitem[\protect\citeauthoryear{Zhao \bgroup \em et al.\egroup
  }{2016}]{Zhao2016}
Ran Zhao, Tanmay Sinha, Alan~W Black, and Justine Cassell.
\newblock Automatic recognition of conversational strategies in the service of
  a socially-aware dialog system.
\newblock In {\em SIGDial}, pages 381--392, 2016.

\bibitem[\protect\citeauthoryear{Zhong \bgroup \em et al.\egroup
  }{2020}]{zhong2020towards}
Peixiang Zhong, Chen Zhang, Hao Wang, Yong Liu, and Chunyan Miao.
\newblock Towards persona-based empathetic conversational models.
\newblock {\em arXiv preprint arXiv:2004.12316}, 2020.

\end{thebibliography}

%\section{Conclusions}

%\pagebreak

%\end{document}

\end{document}